\documentclass[lettersize,journal]{IEEEtran}
\usepackage{amsmath,amsfonts}
\usepackage{algorithmic}
\usepackage{algorithm}
\usepackage{array}
\usepackage[caption=false,font=normalsize,labelfont=sf,textfont=sf]{subfig}
\usepackage{textcomp}
\usepackage{stfloats}
\usepackage{url}
\usepackage{verbatim}
\usepackage{graphicx}
\usepackage{cite}
\usepackage{bm}
\usepackage{amssymb}
\usepackage{amsmath}
\usepackage{array}   
\usepackage{tabularx}
\usepackage{multirow}
\usepackage{orcidlink}
\usepackage{tikz}
\usepackage{amsmath, amsfonts, amsthm, stackrel, amssymb}   
\hypersetup{
colorlinks=true,
linkcolor=black,
citecolor=black
}
\begin{document}

\title{UniDCP: Unifying Multiple Medical Vision-language Tasks via Dynamic Cross-modal Learnable Prompts}

\author{Chenlu Zhan, Yufei Zhang, Yu Lin, Gaoang Wang ~\IEEEmembership{Member,~IEEE}, Hongwei Wang~\IEEEmembership{Senior Member,~IEEE}
\thanks{This work was supported in part by Zhejiang Provincial Natural Science Foundation of China (LDT23F02023F02).}
\thanks{Chenlu Zhan is with the College of Computer Science and Technology, Zhejiang University, Zhejiang, China (chenlu.22@intl.zju.edu.cn)}
\thanks{Yufei Zhang is with College of Biomedical Engineering and Instrument Science, Zhejiang University, Zhejiang, China (yufei1.23@intl.zju.edu.cn)}
\thanks{Yu Lin, Gaoang Wang and Hongwei Wang are with Zhejiang University-University of Illinois Urbana-Champaign Institute, Zhejiang University, Haining, China. (yulin@intl.zju.edu.cn, gaoangwang@intl.zju.edu.cn, hongweiwang@zju.edu.cn)}}

\markboth{Journal of \LaTeX\ Class Files,~Vol.~14, No.~8, December~2023}%
{Shell \MakeLowercase{\textit{et al.}}: A Sample Article Using IEEEtran.cls for IEEE Journals}


\maketitle

\begin{abstract}
Medical vision-language pre-training (Med-VLP) models have recently accelerated the fast-growing medical diagnostics application. However, most Med-VLP models learn task-specific representations independently from scratch, thereby leading to great inflexibility when they work across multiple fine-tuning tasks.
In this work, we  propose \textbf{UniDCP}, a \textbf{Uni}fied 
medical vision-language model with \textbf{D}ynamic \textbf{C}ross-modal learnable \textbf{P}rompts, which can be plastically applied to multiple medical vision-language tasks.
Specifically, 
we explicitly construct a unified framework to harmonize diverse inputs from multiple pre-training tasks by leveraging cross-modal prompts for unification, which accordingly can accommodate heterogeneous medical fine-tuning tasks.
Furthermore, we conceive a dynamic cross-modal prompt optimizing strategy that optimizes the prompts within the shareable space for implicitly processing the shareable clinic knowledge. 
UniDCP is the first Med-VLP model capable of performing all $8$ medical uni-modal and cross-modal tasks over $14$ corresponding datasets, consistently yielding superior results over diverse state-of-the-art methods. 
\end{abstract}


\section{Introduction}
\IEEEPARstart{M}{edical} image analysis with deep learning has significantly enhanced its application in medical practice. However, these medical multi-modal models for medical diagnostic analysis require extensive training on a multitude of diverse medical datasets. This entails time-consuming and complex manual annotation costs, posing a significant impediment to the development of intelligent medical diagnostics. Recently, 
advances in medical vision-language pre-training models (Med-VLP)~\cite{khare2021mmbert,clipeslami2021does,gong2022vqamix,liu2021contrastive,zhang2023semi,zhang2023multi} have significantly improved the 
efficiency of the medical diagnostics tasks. The Med-VLP model aims to learn unified representations from large-scale medical clinic image-text pairs. These representations with task-specific parameters are then fine-tuned for the medical downstream tasks (e.g. medical question answering task, medical report generation task). 
\begin{figure}[htbp]
\centering
\includegraphics[width=1\linewidth]{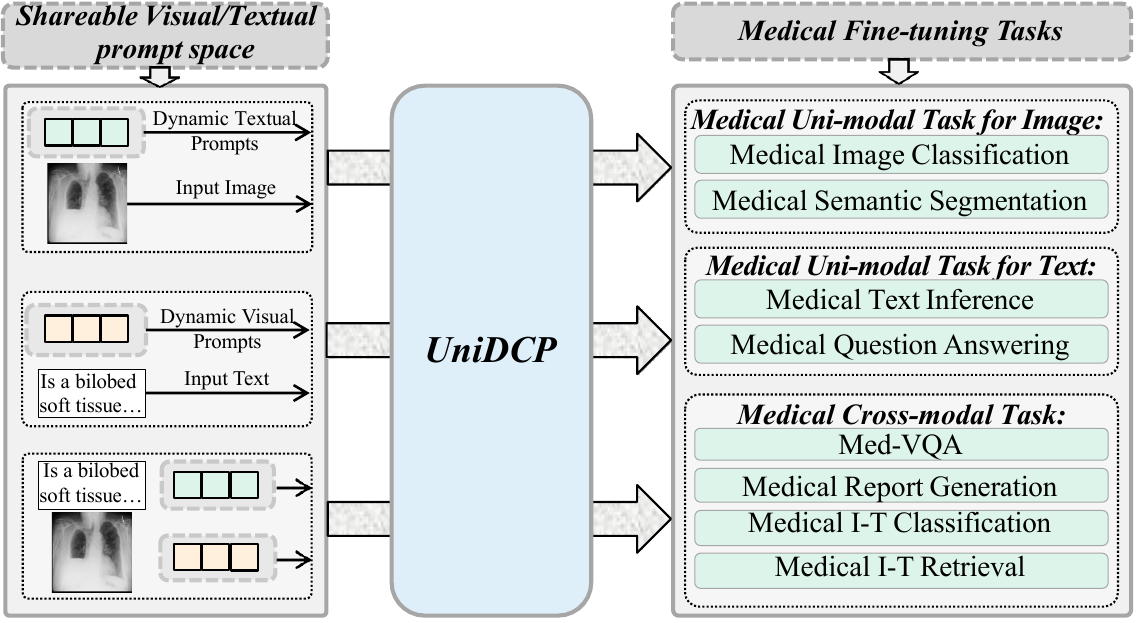}
\caption{UniDCP is a unified and plastic model capable of multiple medical uni-modal and cross-modal tasks with dynamic shareable cross-modal prompts. The broad unification is achieved by harmonizing diverse inputs of heterogeneous tasks with the dynamic visual and textual prompts which are selected from the corresponding shareable space.}
\label{prompt}
\end{figure}

However, most recent Med-VLP works~\cite{m3ae,clipeslami2021does,yan2022clinical,khare2021mmbert,zhang2023multi,zhang2023semi} learn proprietary representations corresponding to fine-tuning tasks, which results in the inflexibility of Med-VLP model across multiple fine-tuning tasks due to task-specific structures or static components.
Equipped with task-specific structures, some works~\cite{huang2021gloria,zhou2022generalized,wu2023medklip} grasp fine-grained representations within two-stream branches for medical classification tasks, and the others~\cite{m3ae,zhou2023advancing} introduce exclusively aligned frameworks for the medical cross-modal reasoning tasks. This trend constrains the transferability of the Med-VLP model with limited concept knowledge.
Besides, some works~\cite{coop,vlp,zang2022unified} introduce the visual and textual prompts into vision-language tasks, which transfer the well-optimized models to fine-tuning tasks. 
Although these works express superior generalization ability in downstream tasks, the learned prompts remain static, 
exclusively benefit the individual fine-tuning task and reflect restrictions for the others.
It remains an open problem to construct a unified and plastic Med-VLP model that can simultaneously adapt to multiple medical tasks.

To address the challenges above, we propose \textbf{UniDCP}, a \textbf{Uni}fied Med-VLP model via \textbf{D}ynamic  \textbf{C}ross-modal learnable \textbf{P}rompts, which is capable of simultaneously performing diverse medical uni-modal and cross-modal fine-tuning tasks.
Specifically, we construct a unified model within a triple-stage prompt scheme: the cross-modal prompt initialization, multi-task pre-training and multi-task adaptation.  As shown in Fig.~\ref{prompt},
the unified model is scaled by harmonizing diverse inputs from multiple pre-training tasks via the initialized cross-modal prompts, which consequently enables the handling of diverse medical fine-tuning tasks with no task-specific parameters.
Furthermore, we initiatively present the dynamic visual-textual prompts optimizing strategy for resiliently cooperating multiple tasks with the shareable clinic knowledge. 
In contrast to static prompts, we encode visual/textual prompts into a key-value shareable space, where dynamic optimization occurs by selecting a subset of the most relevant prompts through the query function.
The key-value setting is devised for {dynamic prompts strategy} which links the prompts as key-value pairs and adopts the query function for semantic prompts selection. It is superior in medical cross-modal knowledge extraction and alignment.
Then we adapt the unified model and the dynamic prompts with shareable clinic knowledge into multiple medical fine-tuning tasks. During these processes, 
{the clinic knowledge is acquired from {two aspects}: {1)} The dynamic prompts with the {most pertinent medical information} are retrieved through a query function. {2)} The selected prompts consecrate with visual/textual representations that {plastically adapt shareable clinic knowledge}  into multiple tasks.}
We conduct comprehensive experiments on medical vision-language understanding benchmarks which conclude $8$ uni-modal/cross-modal fine-tuning tasks and $14$ corresponding medical datasets.
Extensive experiments demonstrate the superiority of the proposed model in plastically integrating clinic-shareable knowledge into diverse medical vision-language tasks within a unified structure.
{Overall, the proposed UniDCP is a {integrated} and {flexible} model with low-cost expansion. 1) Integration: UniDCP has {wider generalization} in various-modality tasks with {no task-specific} modules and achieves {all-sided SOTA} results. 
2) Flexibility: We propose dynamic prompts to select the most semantically effective representations for each task, which {scalably adapts to various tasks with} {negligible overhead}.
Our contributions are summarized as follows:

\begin{itemize}
\item We propose UniDCP, a unified and plastic model capable of multiple medical fine-tuning tasks by harmonizing heterogeneous inputs from multiple pre-training tasks via cross-modal prompts.

\item We present the dynamic cross-modal prompts optimizing strategy within shareable space to cooperate with shareable clinic knowledge under different task transitions.

\item UniDCP is the first Med-VLP model that attains state-of-the-art results on $8$ medical vision-language tasks with $14$ benchmarks, illustrating the superiority in integrating the shareable clinic knowledge into multiple tasks.

\end{itemize}

\section{Related works}

\subsection{Medical Vision-language Pre-training}
Medical vision-language pre-training (Med-VLP) aims to learn the cross-modal representations from large-scale medical image-text pairs. Current Med-VLP methods~\cite{liu2021contrastive,khare2021mmbert,gao2021clip,m3ae,huang2021gloria,wen2023msgfusion,zhang2023multi,zhang2023semi} employ task-specific pre-training strategies to learn proprietary representations, such as the fine-grained representations for classification and the aligned cross-modal representations for reasoning. 
For the former strategy, GLoRIA~\cite{huang2021gloria} presents an efficient medical image recognition method by evaluating the local similarity between image sub-regions and words. REFERS~\cite{zhou2022generalized} enhances medical visual information by adopting multiple visual views. MedKLIP~\cite{wu2023medklip} introduces a triplet extraction module designed to retrieve medical-related details from reports for reinforcing the classification ability.
For the latter, MRM~\cite{zhou2023advancing} acquires the aligned visual knowledge-enhanced representations with the direct incorporation of cross-modal visual information processed by GAP into report reconstruction. M3AE~\cite{m3ae} devises various aligned supervised pre-training tasks for obtaining medical cross-modal domain expertise, which is a very similar setting to ours. These works all focus on learning superior specific representations tailored to corresponding downstream tasks but pose a challenge in consolidating the Med-VLP model for multiple fine-tuning tasks and inhibiting its universal applicability.
There remains a gap in constructing a Med-VLP framework that has superior plasticity and transferability to multiple medical vision-language tasks within a unified model.

\begin{figure*}[htbp]
\centering
\includegraphics[width=1\linewidth]{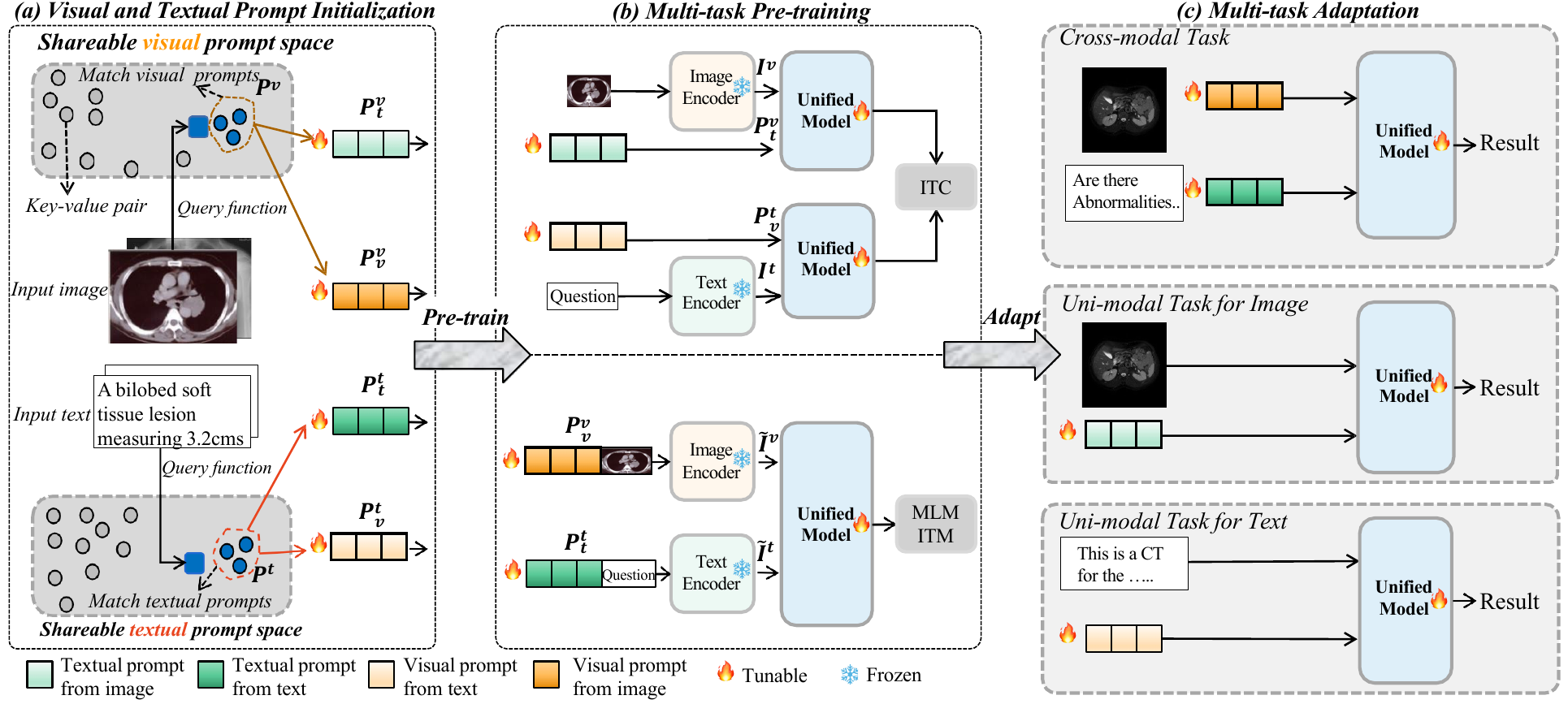}
\caption{The network structure of UniDCP. (a) Within a unified model, the visual and textual prompts are introduced from the shareable visual prompt space to unify the inputs. The cross-modal prompts are initialized by drawing the chosen feature nearer to the respective prompts through the query function. 
(b) After unifying the inputs with the visual/textual prompts, the unified model is optimized with the cross-modal and uni-modal pre-training tasks. (c) The unified model, which incorporates the dynamic visual and textual prompts is adapted to the cross-modal tasks and the uni-modal tasks for text and image.}
\label{fig:UniDCP}
\end{figure*}

\subsection{Prompt Tuning}
The prompt tuning~\cite{zhong2021factual,li2021prefix,lester2021power} is initially introduced by the NLP community to promote the efficiency of the large pre-trained model on fine-tuning language tasks. Recent works~\cite{coop,jia2022visual,gao2021clip,xing2023dual} also explore the prompt tuning in vision-language tasks. CoOp~\cite{coop} designs the learnable prompts with context words and achieves obvious improvements with few labeled images. VPT~\cite{jia2022visual}  turns patches of images into learnable prompts to efficiently fine-tune the large-scale transformer models in vision tasks. UPT~\cite{zang2022unified} introduces two sets of vision-language prompts for fine-tuning the vision-language models. CLIP-Adapter~\cite{gao2021clip} introduces an adapter module to embed the text and image features into the same space for better adaptation of various datasets. Although these static prompt tuning works have superior efficiency in fine-tuning tasks, they restrictively remain static when applied to exclusive downstream tasks independently. It motivates us to explore learnable dynamic cross-modal prompts which can be adapted to various medical fine-tuning tasks with shareable clinic knowledge.
{There are {two key innovations} that distinguish our work from others~\cite{coop,jia2022visual,zang2022unified,chen2023towards}: 1) UniDCP is a {comprehensive yet low-cost} model that seamlessly handles various medical multi-modal and single-modal tasks with dynamic visual/textual prompts while other models with expensive-cost lack the processability of text-only tasks. 2) Besides the multi-modal prediction, classification, segmentation, etc, UniDCP ventures into medical {generation task} and exhibits better performance.}

\section{Methodology}

In this section, we propose UniDCP, which is a unified and plastic model for multiple medical vision-language tasks. Fig. \ref{fig:UniDCP} illustrates the overview of triple-stage procedures which contains the visual-textual prompt initialization, multi-task pre-training and multi-task adaptation.  
\subsection{Problem Setup}


Med-VLP aims to learn medical vision-language representations from clinic images and paired radiology reports. Some works introduce prompts $\emph{{P}}$ that can transfer well-optimized models to fine-tuning tasks. Within the pre-training model $U$, the medical inputs $\emph{{I}}$ may consist of the text-only embedding input ${{I^t}}\in\mathbb{R}^{D_t\times L_t}$,  image-only embedding input ${{I^v}}\in\mathbb{R}^{D_v\times L_v}$  and image-text pairs, where $L_t$, $L_v$ and $D_t$, $D_v$ are the token lengths and the embedding dimension of the textual and visual features. The representations are learned with the pre-trained model from $R$ pre-training tasks, which can be formulated as:
\begin{equation}
\resizebox{0.905\hsize}{!}{$\hat{\mu}, \hat{\mu _1} ,...,\hat{\mu _R},\hat{\theta} = \mathop{\arg\max}\limits_{{\mu},\mu _1 ,...,\mu _R,\theta} \sum\limits_{{\rm{r}} = 1}^R {{{\cal L}_r}\left( {{Y_r},{{\cal H}_{{\mu _r}}}\left( {{U_\mu }\left( \emph{{I}},{P}_{\theta} \right)} \right)} \right)}$ }
\end{equation}
where ${{\cal H}_{{\mu _r}}}$ represents the prediction heads with corresponding trainable parameters $\mu _r$, $Y_r$ is the ground-truth labels and ${\cal L}_r$ is the loss function of the pre-training tasks $r$. $U_\mu$ is the pre-trained model which is parameterized by $\mu$. ${P}_{\theta}$ represents the prompts with tunable parameter $\theta$.

For constructing a unified model $U_{\mu}$ that can apply to multiple medical fine-tuning tasks, the issue lies in enabling the unified model to effectively handle variable inputs and grasp the shareable knowledge.

\subsection{Visual and Textual Prompts Initialization}
To promote compatibility of the model, we unify the inputs $\emph{{I}}$ by introducing the visual prompt ${P^v}$ for the text-only embedding input ${{I^t}}$ and the textual prompt ${P^t}$ for the image-only embedding input ${{I^v}}$, respectively.
\begin{equation}
 \emph{ {{I}}} = \left\{ 
 \begin{array}{l}
({{I^v}},{{P^t}}) \quad $if  image-only$\\
({{P^v}},{{I^t}})  \quad $if text-only$\\
({{I^v}},{{I^t}})\quad $if image-text$\\
\end{array} \right.
\end{equation}
where ${P^t}\in\mathbb{R}^{D_t\times N}$, ${P^v}\in\mathbb{R}^{D_v\times N}$ and $N$ represents the number of the learnable prompts.

Moreover, in contrast to static prompts tailored for each independent medical downstream task, we introduce dynamic cross-modal prompts that plastically incorporate shareable clinic visual/textual knowledge into multiple medical downstream tasks.
Formally, we first construct the shareable visual and textual prompt space $P^V$, $P^T$ to store the transferable clinic prompt knowledge. 
{The proposed shareable prompt space exactly serves as a {cross-modal representation repository} that adapts the most effective representations for various fine-tuning tasks through the dynamic strategy, 
which can be defined as:
\begin{equation}
\begin{aligned}
\label{mn}
    & P^V=\{{{p_1^v},{p_2^v},...,{p_M^v}}\} \\
    & P^T=\{{{p_1^t},{p_2^t},...,{p_J^t}}\} \\
\end{aligned}
\end{equation}
where $M$, $J$ is the total number of the visual and textual prompts in each space respectively, ${p_i^v}\in\mathbb{R}^{D_v\times L_v^p}$, ${p_j^t}\in\mathbb{R}^{D_t\times L_t^p}$ are the single visual and textual prompt with the token length $L_v^p$, $L_t^p$, and embedding sizes $D_v$, $D_t$ are same as the corresponding inputs.

Following the VQ-VAE~\cite{van2017neuralVQ-VAE}, we further devise the cross-modal prompts query strategy for selecting the most effective $N$ prompts from the corresponding shareable prompt space, applying to unify the inputs. Specifically, we link each visual/textual prompt as the trainable key-value pair: $\{({k_1^v},{p_1^v}),...,({k_M^v}, {p_M^v})\}$ and $\{({k_1^t},{p_1^t}),...,({k_J^t}, {p_J^t})\}$, where the ${k_i^t}\in\mathbb{R}^{D_t}$ and the ${k_j^v}\in\mathbb{R}^{D_v}$.
Thus we adopt the query function $q$ to encode the visual/textual feature into the same dimension $\mathbb{R}^{D_t}$ and $\mathbb{R}^{D_v}$ as the key in the key-value pair. With the visual and textual embedding inputs ${{I^v}}$ and ${{I^t}}$, we query the top-N keys with the query function $q$:
\begin{equation}
    \begin{aligned}
    \label{top-NN}
       & {P^t}={\mathop{\arg\max}\limits_{{\left\{s_i\right\}_{i=1}^N}\subseteq\left[1,M\right]}{\sum_{i=1}^{N}{\gamma\left(q({I^t}\right),{k_{s_i}^t})}} }\\
       & {P^v}={\mathop{\arg\max}\limits_{{\left\{s_i\right\}_{i=1}^N}\subseteq\left[1,J\right]}{\sum_{i=1}^{N}{\gamma\left(q({I^v}\right),{k_{s_i}^v})}} }\\
    \end{aligned}
\end{equation}
where the $\{{s_i}\}_{i=1}^N$ is the subset of length $N$, the $\gamma$ represents the matching function that computes the cosine similarity between the key-value prompt and the query to
select effective prompts.
Overall, the surrogate loss $L_p$ for drawing the chosen keys nearer to their respective query features can be denoted:
\begin{equation}
    L_p=\sum_{{P^t}}{\gamma\left(q({I^t}\right),{k_{s_i}^t})+}\sum_{{P^v}}{\gamma\left(q({I^v}\right),{k_{s_i}^v})}
\end{equation}



\subsection{Multi-task Pre-training}
To accommodate diverse pre-training tasks with various inputs, we treat the visual prompt ${P^v}$ as ${P_t^v}$ and ${P_v^v}$, the textual prompt ${P^t}$ as ${P_t^t}$ and ${P_v^t}$, playing distinct roles related to vision and text when the text-only and image-only input, respectively. 
We can both unify various inputs and adapt cross-modal inputs into ${\widetilde{I}^v}$ and ${\widetilde{I}^t}$ by concatenating the inputs and the prompts ${P_v^v}$, ${P_t^t}$ along $D_v$ and $D_t$ dimension respectively. This allows us to optimize the same unified model with multiple pre-training tasks, as illustrated in Fig. \ref{fig:UniDCP} (b). 
Following the VisualBert~\cite{li2019visualbert}, we employ the masked language modeling (MLM),  image-text matching (ITM) and image-text contrast (ITC) tasks for optimizing the unified model $U$. 
For MLM and ITM, dynamic prompts are also concatenated with image/text embeddings to 
facilitate cross-modal alignments.
\subsubsection{MLM}
The MLM pre-train task randomly masks 15\% input words ${Y_{MLM}}$ and the optimization object across the inputs ${{\widetilde{I}^v}}$ and ${{\widetilde{I}^t}}$ is:
\begin{equation}
    {L_{MLM}} =  - \sum\limits_{({{\widetilde{I}^v}},{{\widetilde{I}^t}})} {\log {f_s}({H_{MLM}})} ({Y_{MLM}}|{{\widetilde{I}^v}},{\widetilde{I}^t_M})
\end{equation}
where $H_{MLM}$ represents prediction heads of MLM, $f_s$ is the softmax layer, ${\widetilde{I}^t_M}$ is the remaining text input.
\subsubsection{ITM}The ITM task randomly selects positive and negative image-text pairs into the encoder,
proposing to figure out the matching degree of the input image and text features. The final probability $p_{ITM}$ can be calculated by the output of the softmax layer with the concatenation of the representations of the image-text pairs 
$z_{[CLS]}^v$ and $z_{[CLS]}^t$. The optimization objective of ITM is defined as:
\begin{equation}
    {L_{ITM}} =  - \sum\limits_{({{\widetilde{I}^v}},{{\widetilde{I}^t}})} {\log p_{ITM}} ({Y_{ITM}}|{{\widetilde{I}^v}},{\widetilde{I}^t})
\end{equation}
\subsubsection{ITC}
The ITC task adopts the dual-encoder to explore the best cross-modal representation through computing the similarity between image-to-text $p_n^{i2t}$ and text-to-image $p_n^{t2i}$. 
The optimization objective of ITC is defined as:
\begin{equation}
     \begin{aligned}
    & {L_{ITC}} =  - \frac{1}{2}\sum\limits_{{({{I}^v}},{P_t^v})} {\log {p^{i2t}}({Y^{i2t}}|{{I}^v},{P_t^v})}  \\
    & - \frac{1}{2}\sum\limits_{({{I}^t},{P_v^t})} {\log {p^{t2i}}({Y^{t2i}}|{{I}^t},{P_v^t})}
     \end{aligned}
\end{equation}
where $Y^{i2t}$, $Y^{t2i}$ are ground-truth labels which set to $0$ for the negative pair and $1$ for the positive pair.
Overall, the total optimization $L_{pre}$ of the multi-task pre-training consists of three loss terms:
\begin{equation}
    \label{loss}
    {L_{pre}} = {L_{MLM}} +  \sigma {L_{ITM}} + \lambda {L_{ITC}} +\beta  {L_p}
\end{equation}
where $\sigma$, $\lambda$, $\beta$ are hyperparameters for balancing the loss terms.

\subsection{Multi-task Adaptation}
As shown in Fig.\ref{fig:UniDCP} (c),
leveraging a unified model that incorporates the dynamic visual and textual prompts, we apply it to cross-modal and uni-modal medical vision-language fine-tuning benchmarks. For the report generation, our unified model extracts image representations and then combines them with text prompts. The concatenations are fed into a transformer decoder (except cross-attention layers) initialized from a pre-trained language encoder. 

Specifically, initialized from the pre-trained model $U$, we utilize the combined representations of image/visual prompt ${\widetilde{I}^v}$ and text/textual prompt ${\widetilde{I}^t}$ as input for prediction in the medical cross-modal task. For the medical uni-modal task for image/text, we unify the input with the contrary dynamic prompt and then feed the sequence $({{I^v}},{{P^v_t}})$/$({{P^t_v}},{{{I^t}}})$ to the unified model with initialized parameters.

\section{Datasets and Implementation Details}
\subsection{Pre-train Datasets and Configurations}
We pre-train the proposed framework with the ROCO~\cite{pelka2018radiology} and MIMIC-CXR~\cite{johnson2019mimic} datasets.
The ROCO~\cite{pelka2018radiology} is the large-scare public medical image-text dataset that includes $81,000$ radiology images of various modalities. The MIMIC-CXR~\cite{johnson2019mimic} is a large X-ray dataset that concludes $377,100$ radiology images of the chest and $227,835$ corresponding reports from patients.
We all follow the official splitting.
\subsection{Medical Fine-tuning Datasets}
\label{A}
We conduct medical vision-language benchmarks which conclude $8$ medical cross-modal and uni-modal tasks with $14$ fine-tuning medical datasets. 
\subsubsection{Medical Vision Question Answering}
  {VQA-RAD}~\cite{vqaradlau2018dataset}
comprises $315$ radiology images sourced from PubMed, alongside $3,515$ question-and-answer pairs meticulously curated by clinicians. We follow the original splitting which contains $3064$ training samples and $451$ samples for testing.
{SLAKE}~\cite{slakeliu2021slake}
concludes $642$ radiology images and $7,033$ question-and-answer samples across $39$ organs. We adopt the official splitting of $70\%$ training pairs and $30\%$ testing pairs.
{Path-VQA}~\cite{pathvqahe2021towards} is a pathology dataset for Med-VQA which contains $4,998$ pathology images with $32,795$ question-answer pairs. The image-question pairs are composed of $8$ types of questions.

 \subsubsection{Medical Report Generation}
{IU X-Ray}~\cite{demner2016preparing} is the predominant dataset employed for the medical report generation task, which contains $7,470$ chest X-ray images and $3,955$ related clinic reports.
The MIMIC-CXR~\cite{johnson2019mimic} dataset encompasses $377,100$ radiology images of the chest, along with $227,835$ corresponding reports from patients. We use the train-set of MIMIC-CRX for fine-tuning and the test-set for evaluation.
\subsubsection{Medical Image-Text Classification}
{MELINDA}~\cite{wu2021melinda}
 is a multi-modal biomedical dataset that includes $2833$ medical images paired with corresponding captions. We follow the official data splitting with $80\%$-$10\%$-$10\%$ of train-set, validation-set and test-set respectively.
 \subsubsection{Medical Image-Text Retrieval}
  We filter the non-radiology samples of {ROCO}~\cite{pelka2018radiology} and split the $65,460$ medical image-caption pairs as the train-set, $8,183$ and $8,182$ samples as the validation-set and test-set respectively.

\begin{table*}
	\centering
           \caption{The comparisons between UniDCP and SOTA Med-VQA methods. ``Open": open-ended questions with no fixed form answer. ``Closed": closed-ended questions with a yes/no answer.-PMC-VQA: Pre-train on the PMC-VQA dataset. } 
	\scalebox{0.85}{
\small
	\setlength{\tabcolsep}{9pt}
\begin{tabular}{llllllllll}
\hline
\multirow{2}{*}{Methods} & \multicolumn{3}{c}{VQA-RAD(\%)} & \multicolumn{3}{c}{SLAKE(\%)} & \multicolumn{3}{c}{Path-VQA(\%)} \\ \cline{2-10} 
                         & Open   & Closed    & All    & Open   & Closed  & All    & Open    & Closed   & All    \\ \hline
MFB~\cite{mfbyu2017multi}                      & 14.5   & 74.3      & 50.6   & 72.2   & 75.0      & 73.3   & -       & -        & -      \\
SAN~\cite{sanyang2016stacked}                      & 31.3   & 69.5      & 54.3   & 74.0     & 79.1    & 76.0     & 1.6     & 59.4     & 30.5   \\
BAN~\cite{kim2018bilinear}                      & 37.4   & 72.1  & 58.3   & 74.6   & 79.1    & 76.3   & 2.9     & 68.2     & 35.2   \\ \hline
MAML+SAN~\cite{mamlffinn2017model}                 & 38.2   & 69.7      & 57.1   & -      & -       & -      & 5.4     & 75.3     & 40.5   \\
MAML+BAN~\cite{mamlffinn2017model}                 & 40.1   & 72.4      & 60.7   & -      & -       & -      & 5.9     & 79.5     & 42.9   \\
MEVF+SAN~\cite{mevf_san_nguyen2019overcoming}                 & 49.2   & 73.9      & 64.1   & 75.3   & 78.4    & 76.5   & 6.0     & 81.0     & 43.6   \\
MEVF+BAN~\cite{mevf_san_nguyen2019overcoming}                 & 49.2   & 77.2      & 66.1   & 77.8   & 79.8    & 78.6   & 8.1     & 81.4     & 44.8   \\
VQAMIX~\cite{gong2022vqamix}                   & 56.6   & 79.6      & 70.4   & -      & -       & -      & 12.1    & 84.4     & 48.4   \\
MMBERT~\cite{khare2021mmbert}                   & 58.3   & 76.9      & 66.9   & -      & -       & -      & -       & -        & -      \\ 
CPRD+BAN~\cite{liu2021contrastive}                 & 52.5   & 77.9      & 67.8   & 79.5   & 83.4    & 81.1   & -       & -        & -      \\
MedVInT~\cite{zhang2023pmc}                     & 55.3  & 80.5   & 70.5  & 79.7  & 85.1   & 81.8  & -  & -      & - \\
PubMedCLIP~\cite{clipeslami2021does}               & 58.0   & 79.6      & 71.1  & 78.2   & 82.6   & 80.1   & 13.4       & 83.6      & 48.5      \\
M3AE~\cite{m3ae}                     & 67.2  & 83.5   & 77.0  & 80.3  & 87.8   & 83.2  & 14.2   & 84.0      & 48.8   \\ \hline
\textbf{Ours }                    & \textbf{68.5$_{\pm 1.2}$  }   & \textbf{84.4$_{\pm 1.0}$   }     & \textbf{79.2$_{\pm 1.1}$  }   & \textbf{82.4$_{\pm 1.4}$}    & \textbf{88.9$_{\pm 1.2}$}    & \textbf{85.9$_{\pm 1.3}$}   & \textbf{15.9$_{\pm 1.4}$ }   & \textbf{85.4$_{\pm 1.3}$}     & \textbf{50.7$_{\pm 1.4}$} \\ \hline
MedVInT-PMC-VQA~\cite{zhang2023pmc}                     & 73.7  & 86.8   & 81.6  & 84.5  & 86.3   & 85.2  & -   & -      & -   \\
\textbf{Ours-PMC-VQA}                   & \textbf{74.5$_{\pm 1.1}$}              & \textbf{87.9$_{\pm 1.0}$}              & \textbf{82.7$_{\pm 1.1}$}    & \textbf{85.8$_{\pm 1.1}$}    & \textbf{88.6$_{\pm 1.4}$}    & \textbf{86.7$_{\pm 1.2}$}   & \textbf{16.9$_{\pm 1.2}$ }   & \textbf{86.2$_{\pm 1.4}$}     & \textbf{51.9$_{\pm 1.3}$} \\ \hline
\end{tabular}
}
	\label{mvqa}  
\end{table*}
\subsubsection{Medical Image Classification}

{CheXpert}~\cite{irvin2019chexpert}
   is a large radiograph dataset with $224,316$ radiographs. Since the private access of the test-set, following ConVIRT~\cite{convirt}, we adopt the randomly selected $5000$ radiographs of the train-set as the validation-set and take the validation-set as the test-set.
{RSNA Pneumonia}~\cite{shih2019augmenting}
collects nearly $30,000$ frontal chest X-rays. We split the dataset into $25184$, $1500$ and $3,000$ of train-set, validation-set and test-set respectively.

\subsubsection{Medical Semantic Segmentation}
The SIIM Pneumothorax dataset~\cite{siimAimoldin2019} encompasses $12,047$ chest radiographs. The dataset is partitioned into training, validation, and testing subsets, accounting for 70\%, 30\%, and 30\% respectively.
The RSNA~\cite{rsnashih2019augmenting} dataset follows the same split protocol. We transform the ground truth of object detection into masks suitable for semantic segmentation tasks.

\subsubsection{Medical Natural Language Inference}
{MedNLI}~\cite{mednli}
is adopted for the clinic natural language inference task. We follow the original splitting which contains $11,232$ training samples, $1,395$ development samples, and $1,422$ test samples.




\subsubsection{Medical Question Answering}
{PubMedQA}~\cite{jin2019pubmedqa} is a biomedical question answering dataset selected from PubMed abstracts. PubMedQA consists of $1,000$ instances expertly labeled, along with $61,200$ unlabeled and $211,300$ question-answer instances artificially generated. We select $450$, $50$, $500$ question-answer instances for the train-set, validation-set, and test-set respectively.
{BioAsq}~\cite{bioasqnentidis2020results}.
The BioAsq is a biomedical dataset for the medical question answering task, which contains $618$ training samples and $161$ testing samples.


\subsection{Implementation Details}

UniDCP is trained with $6$ NVIDIA $3090$ GPUs.  For pre-training, we adopt 12-layer transformer initialized from CLIP-ViT~\cite{radford2021learning} and RoBERTa~\cite{liu2019roberta} as the vision and text encoder. The $6$-layer transformer with $768$ dimensions of hidden states and $12$ heads is settled as the unified model. The model pre-trains $100,000$ steps with AdamW optimizer whose learning rate of vision and text encoder are $1{{e}^{-5}}$ and $5{{e}^{-5}}$ respectively, and the weight decay is $5{{e}^{-4}}$. 
We conduct the effect results of transformer layers in Fig.~\ref{layer}. 
The input images are cropped to $224\times224$. The hyperparameters $\lambda$, $\beta$ and $\sigma$ of pre-training loss are set to $0.8$, $0.9$ and $0.9$ respectively, the sensitivity results are in Fig.~\ref{layer} and \ref{hyper}. 
The dimensions $D_v$ and $D_t$ of visual and textual embedding are $1024$ and $768$ respectively. The top-N in Eqn. \ref{top-NN} is set to top-$5$.  The final $M$, $J$ are all set to $1024$.
The total number $N$ of learnable visual/textual prompts are set to $49$ and $32$ respectively.

Except for the accuracy of Med-VQA and medical text tasks, we adopt the commonly used NLG metrics for medical report generation, Dice score for medical semantic segmentation and AUC score for medical image classification.

\section{Experiments and Results}

\subsection{Comparison with State-of-the-art Methods}
To demonstrate the effectiveness of UniDCP, we conduct abundant experiments on $8$ medical cross-modal and uni-modal tasks over $14$ datasets in total.

\begin{table*}
	\centering
 \caption{The comparisons between UniDCP and medical report generation methods on IU X-Ray and MIMIC-CXR datasets.} 
	\scalebox{0.85}{
\small
	\setlength{\tabcolsep}{3pt}
	
\begin{tabular}{lllllllllll}
\hline
\multirow{2}{*}{Methods} & \multicolumn{5}{c}{\begin{tabular}[c]{@{}c@{}}IU X-Ray(mean${\pm}$std)\end{tabular}}                        & \multicolumn{5}{c}{MIMIC-CXR(mean${\pm}$std)}                                                                       \\ \cline{2-11} 
                         & BLEU-1         & BLEU-2         & BLEU-3         & BLEU-4         
                         & ROUGE-L       & BLEU-1         & BLEU-2         & BLEU-3         & BLEU-4         
                         & ROUGE-L       \\ \hline
Show-Tell~\cite{vinyals2015show}
& 0.346          & 0.214          & 0.141          & 0.095          
& 0.320           & 0.299          & 0.184          & 0.121          & 0.084          
& 0.263          \\
Att2in~\cite{rennie2017self}
& 0.399          & 0.239          & 0.172          & 0.126          
& 0.321          & 0.325          & 0.203          & 0.136          & 0.096          
& 0.276          \\
Transformer~\cite{vaswani2017attention}
& 0.422          & 0.264          & 0.177          & 0.120           
& 0.338          & 0.314          & 0.192          & 0.127          & 0.090           
& 0.265          \\
R2Gen~\cite{chen2020generating}
& 0.470          & 0.304          & 0.219          & 0.165          
& 0.371          & 0.353          & 0.218          & 0.145          & 0.103          
& 0.277          \\
R2GenCMN~\cite{chen2022cross}
& 0.475          & 0.309          & 0.222          & 0.170           
& 0.375          & 0.353          & 0.218          & 0.148          & 0.106          
& 0.278          \\
PPKED~\cite{liu2021exploring}
& 0.483          & 0.315          & 0.224          & 0.168          
& 0.376          & 0.360          & 0.224          & 0.149          & 0.106          
& 0.284          \\
AlignTrans~\cite{you2021aligntransformer}
& 0.484          & 0.313          & 0.225          & 0.173          
& 0.379          & 0.378          & 0.235          & 0.156          & 0.112          
& 0.283          \\
Clinical-BERT~\cite{yan2022clinical}
& 0.495          & 0.330          & 0.231          & 0.170          
& 0.376          & 0.383          & 0.230          & 0.151            & 0.106            
& 0.275          \\ 
RAMT~\cite{zhang2023semi}
& 0.482$_{\pm 0.009}$          & 0.310$_{\pm 0.006}$          & 0.221$_{\pm 0.004}$          & 0.165$_{\pm 0.003}$          
& 0.377$_{\pm 0.003}$           & 0.362$_{\pm 0.007}$          & 0.229$_{\pm 0.006}$          & 0.157$_{\pm 0.003}$            & 0.113$_{\pm 0.002}$            
& 0.289$_{\pm 0.003}$          \\ 
TriNet~\cite{yang2021jointmrg}
& 0.478         & 0.344         & 0.248       & 0.180         
& 0.398         & 0.362         & 0.251        & 0.188            & 0.143
& 0.326        \\ 
M2KT~\cite{yang2023radiologym2kt}
& 0.497 &0.319 &0.230& 0.174    &0.399     
& 0.386 &0.237 &0.157 &0.111 &0.274  \\   
MPMA~\cite{zhang2023multi}
& 0.518$_{\pm 0.005}$ &0.337$_{\pm 0.005}$ &0.253$_{\pm 0.003}$& 0.179$_{\pm 0.001}$    &0.388$_{\pm 0.006}$     
& 0.392$_{\pm 0.004}$ &0.246$_{\pm 0.004}$ &0.166$_{\pm 0.002}$ &0.122$_{\pm 0.003}$ &0.295$_{\pm 0.002}$  \\ \hline 
\textbf{Ours}                     & \textbf{0.527}$_{\pm 0.009}$ & \textbf{0.349$_{\pm 0.010}$} & \textbf{0.267$_{\pm 0.011}$} & \textbf{0.195$_{\pm 0.009}$}  
& \textbf{0.421$_{\pm 0.008}$} & \textbf{0.416$_{\pm 0.007}$} & \textbf{0.271$_{\pm 0.009}$} & \textbf{0.199$_{\pm 0.011}$} & \textbf{0.148$_{\pm 0.010}$} 
& \textbf{0.335$_{\pm 0.009}$} \\ \hline
\end{tabular}}
	\label{report generation}
\end{table*}

\begin{table*}

	\centering
 \caption{The experiments on the medical image classification task and medical semantic segmentation task. *: Re-implement result. -C: Pre-trained on CheXpert dataset. -M: Pre-trained on MIMIC-CXR dataset. We conducted the mean accuracy and standard deviation by 5 runs under 5 different seeds. $\dag$: The {70/15/15\%} train/val/test splitting and settings of SIIM dataset.}  
      \setlength{\tabcolsep}{5pt}

  \small
	\scalebox{0.85}{

\begin{tabular}{lllllllllllll}
\hline
\multirow{3}{*}{Methods} & \multicolumn{6}{c}{Medical Image Classification}                                              & \multicolumn{6}{c}{Medical Semantic Segmentation(Dice)}                                     \\ \cline{2-13} 
                         & \multicolumn{3}{c}{CheXpert}                  & \multicolumn{3}{c}{RSNA Pneumonia}            & \multicolumn{3}{c}{SIIM}                      & \multicolumn{3}{c}{RSNA}                      \\ \cline{2-13}
                                                                                    & 1\%           & 10\%          & 100\%         & 1\%           & 10\%          & 100\%         & 1\%           & 10\%          & 100\%         & 1\%           & 10\%          & 100\%         \\ \hline
ConVIRT-C~\cite{convirt}                   
& 85.9          & 86.8          & 87.3          & 77.4          & 80.1          & 81.3          & 25.0          & 43.2          & 59.9          & 55.0          & 67.4          & 67.5          \\
GLoRIA-C~\cite{huang2021gloria}                   
& 86.6          & 87.8          & 88.1          & 86.1          & 88.0          & 88.6          & 35.8          & 46.9          & 63.4          & 59.3          & 67.5          & 67.8          \\
ConVIRT-M~\cite{convirt}                  
& 87.0          & 88.1          & 88.1          & 88.8          & 91.5          & 92.7          & -             & -             & {-}    & {-}    & {-}    & -             \\
MedKLIP-M~\cite{wu2023medklip}                   
& -             & -             & {-}    & 87.3          & 88.0          & 89.4          & 50.2          & 60.8          & 63.9          & 66.2          & 69.4          & 71.9          \\
GLoRIA-M~\cite{huang2021gloria}                   
& 86.5          & 87.5          & 87.8          & 89.7          & 91.2          & 92.1          & 37.4          & 57.1          & 64.0          & 60.3          & 68.7          & 68.3          \\
REFERS-M~\cite{zhou2022generalized}                  
& 87.2          & 88.1          & 88.2          & 89.4          & 91.6          & 92.7          & -             & -             & {-}    & {-}    & {-}    & -             \\
M3AE*-M~\cite{m3ae}                     
& 86.2          & 87.3          & 87.9          & 89.0          & 90.8          & 92.3          & 52.3          & 61.2          & 64.0          & 67.8          & 70.2          & 72.6          \\
MRM*-M~\cite{zhou2023advancing}                      
& 88.5$_{\pm 0.7}$           & 88.5$_{\pm 0.6}$           & 88.7$_{\pm 0.3}$       & 91.3$_{\pm 0.6}$           & 92.7$_{\pm 0.4}$           & 93.3$_{\pm 0.4}$           & 51.4$_{\pm 0.8}$             & 62.3$_{\pm 1.1}$             & 64.8$_{\pm 0.9}$    & 68.5$_{\pm 0.9}$    & 71.3$_{\pm 0.6}$    & 74.7$_{\pm 0.8}$          \\
CheXzero-M~\cite{tiu2022expertCheXzero}                      
& -         & -         & 88.9         & -          & -       & -         & -             & -             & {-}    & {-}    & {-}    & -    \\   
KAD-M~\cite{zhang2023knowledgeKAD}                      
& -         & -         & 90.5          & -        & -         & -          & -             & -             & {-}    & {-}    & {-}    & -       \\ 
MPMA-M~\cite{zhang2023multi}                      
& 89.1$_{\pm 0.8}$         &89.8$_{\pm 0.5}$         & 90.6$_{\pm 0.4}$         & 91.3$_{\pm 0.6}$        & 93.4$_{\pm 0.5}$         & 94.1$_{\pm 0.3}$          & -             & -             & {-}    & {-}    & {-}    & -       \\ \hline
\textbf{Ours-M}            
& \textbf{90.5}$_{\pm 0.9}$ & \textbf{91.2}$_{\pm 0.6}$ & \textbf{91.9}$_{\pm 1.0}$ & \textbf{93.1}$_{\pm 0.7}$  & \textbf{94.6}$_{\pm 0.8}$ & \textbf{95.2}$_{\pm 1.1}$ & \textbf{57.8}$_{\pm 0.8}$ & \textbf{63.4}$_{\pm 0.9}$ & \textbf{65.5}$_{\pm 0.9}$ & \textbf{73.8}$_{\pm 0.7}$ & \textbf{75.4}$_{\pm 0.9}$ & \textbf{77.9}$_{\pm 1.0}$ \\ \hline
MRM$\dag$-M~\cite{zhou2023advancing}                      
& 88.5$_{\pm 0.7}$           & 88.5$_{\pm 0.6}$           & 88.7$_{\pm 0.3}$          & 91.3$_{\pm 0.6}$           & 92.7$_{\pm 0.4}$           & 93.3$_{\pm 0.4}$             & -              & 73.2$_{\pm 0.5}$                     & 91.4$_{\pm 0.3}$     & 68.5$_{\pm 0.9}$    & 71.3$_{\pm 0.6}$    & 74.7$_{\pm 0.8}$         \\
\textbf{Ours$\dag$-M}            
& \textbf{90.5}$_{\pm 0.9}$ & \textbf{91.2}$_{\pm 0.6}$ & \textbf{91.9}$_{\pm 1.0}$ & \textbf{93.1}$_{\pm 0.7}$  & \textbf{94.6}$_{\pm 0.8}$ & \textbf{95.2}$_{\pm 1.1}$ & - & \textbf{74.5}$_{\pm 0.7}$ & \textbf{92.4}$_{\pm 0.6}$& \textbf{73.8}$_{\pm 0.7}$ & \textbf{75.4}$_{\pm 0.9}$ & \textbf{77.9}$_{\pm 1.0}$  \\ \hline
\end{tabular}}
	\label{mic}
\end{table*}

\subsubsection{Medical Visual Question Answering}
   We take comparison experiments on VQA-RAD~\cite{vqaradlau2018dataset}, SLAKE~\cite{slakeliu2021slake} and Path-VQA~\cite{pathvqahe2021towards} datasets in Table~\ref{mvqa}. The proposed UniDCP demonstrates superior performance and obtains the best accuracy compared with SOTA models, including the advanced attention baselines like MFB~\cite{mfbyu2017multi}, SAN~\cite{sanyang2016stacked}, BAN~\cite{kim2018bilinear}, MAML~\cite{mamlffinn2017model} and MEVF~\cite{mevf_san_nguyen2019overcoming}.
\begin{table}[!htbp]
	\centering
 \caption{The experiments on medical image-text classification and retrieval tasks over the MELINDL and ROCO test set.} 
      \setlength{\tabcolsep}{1pt}

	\scalebox{0.75}{
\small
\begin{tabular}{llllllll}
\hline
                                   & {MELINDA}                           & \multicolumn{6}{c}{{ROCO(\%)}}                                                                                                                                                                                 \\ \cline{2-8} 
                                   &                                            & \multicolumn{3}{c}{{T2I}}                                                                        & \multicolumn{3}{c}{{I2T}}                                                                        \\ \cline{3-8} 
\multirow{-3}{*}{{Methods}} & \multirow{-2}{*}{{Test  Acc}} & { { R@1}} & { { R@5}} & {{ R@10}} & {{ R@1}} & { { R@5}} & {{ R@10}} \\ \hline
ViLT~\cite{kim2021vilt}    
& {-}                                          & 9.75 & 28.95                            & 41.40                             & 11.90                            & 31.90                           & 43.20                             \\
METER~\cite{dou2022empirical}  
& -                                          & 11.30                            & 27.25                            & 39.60                             & 14.45                            & 33.30                            & 45.10                             \\
CPRD+BAN~\cite{liu2021contrastive}   
& 75.7$_{\pm 0.8}$                                       & 15.40$_{\pm 0.9}$                            & 35.87$_{\pm 0.8}$                            & 43.45$_{\pm 1.0}$                             & 17.12$_{\pm 1.1}$                            & 38.26$_{\pm 0.9}$                            & 54.76$_{\pm 1.0}$                             \\
PubMedCLIP~\cite{clipeslami2021does}  
& 76.3$_{\pm 0.8}$                                       & 16.92$_{\pm 1.1}$                            & 38.89$_{\pm 0.8}$                            & 49.31$_{\pm 1.1}$                             & 18.02$_{\pm 0.9}$                            & 40.58$_{\pm 1.0}$                           & 56.43$_{\pm 1.2}$                             \\
M3AE~\cite{m3ae}   
& 78.5
& 19.05
& 47.75
& 61.35
& 19.10
& 45.60
& 61.20
\\ \hline
\textbf{Ours}                      & \textbf{79.6}$_{\pm0.5}$                              & \textbf{22.40}$_{\pm 0.6}$                   & \textbf{53.10}$_{\pm 0.9}$                   & \textbf{66.90}$_{\pm 1.0}$                     & \textbf{23.40}$_{\pm 0.8}$                   & \textbf{51.30}$_{\pm 0.9}$                   & \textbf{66.00}$_{\pm 0.9}$                     \\ \hline
\end{tabular}}
	\label{cross-modal1}
\end{table}
   Compared to advanced methods~\cite{khare2021mmbert,liu2021contrastive,clipeslami2021does}
pre-trained with the same medical datasets, our model outperforms by up to $12.3\%$, $8.1\%$ and $11.4\%$ overall accuracy on VQA-RAD respectively, indicating that UniDCP learns predominant unified medical cross-modal representations. Moreover, the improvements over the M3AE~\cite{m3ae} which also pre-trains with multiple tasks on clinic datasets are $2.2\%$, $2.7\%$ and $1.9\%$ overall accuracy on VQA-RAD, SLAKE and Path-VQA respectively. The superior promotion demonstrates the efficacy of dynamic prompts in knowledge alignments. Besides, we observe that UniDCP can considerably exceed the VQAMix~\cite{gong2022vqamix} which devises a special method for better adaptation. When compared with MedVInT~\cite{zhang2023pmc} which pre-trained on the PMC-VQA dataset, our model demonstrates superior performance, achieving improvements of over $1.1\%$ on the VQA-RAD dataset and $1.5\%$ on the SLAKE dataset,  indicating the superior generalization.
  
\subsubsection{Medical Report Generation}
    We adopt the IU X-Ray and MIMIC-CXR (test-set) for evaluating the similarity scores (e.g. BLUE, BOUGE-L) between the generated and labeled reports. As shown in Table \ref{report generation}, our model is superior to the attention works Show-Tell~\cite{vinyals2015show}, Att2in~\cite{vaswani2017attention} and Transformer~\cite{rennie2017self}, and achieves $0.421$ and $0.335$ on the ROUGE-L
    of IU X-Ray and MIMIC-CXR respectively. Specifically, the improvements over the advanced works R2Gen~\cite{chen2020generating}, R2GenCMN~\cite{chen2022cross}, PPKED~\cite{liu2021exploring}, AlignTrans~\cite{you2021aligntransformer} without pre-training are $0.030$, $0.025$, $0.027$, $0.022$ and $0.045$, $0.042$, $0.042$, $0.036$ of the BLUE-4 on two datasets respectively.
    Also, our model outperforms the SOTA works Clinical-BERT~\cite{yan2022clinical} which pre-trains on the domain-specific clinic datasets, M2KT~\cite{yang2023radiologym2kt} which is equipped with the knowledge graph, RAMT~\cite{zhang2023semi}, TriNet~\cite{yang2021jointmrg} and MPMA~\cite{zhang2023multi} by up to $0.060$, $0.061$, $0.046$, $0.009$ and $0.040$ on the ROUGE-L of MIMIC-CXR dataset, respectively. A possible explanation could be that the learned visual and textual prompts benefit the cognizant of the medical images and thus further generate more comprehensive clinic reports. 

\begin{table}
	\centering
 \caption{The experiments on medical natural language inference(NLI) and question answering task. *: Re-implement.}  
      \setlength{\tabcolsep}{6pt}
	
	\scalebox{0.85}{
\small
\begin{tabular}{llll}
\hline
\multicolumn{1}{c}{\multirow{2}{*}{{Methods}}} & {NLI} & \multicolumn{2}{c}{Question Answering} \\ \cline{2-4} 
\multicolumn{1}{c}{}                                  & {MedNLI(\%)}    & PubMedQA(\%)           & BioAsq(\%)            \\ \hline
BioBERT~\cite{lee2020biobert}
& 82.63              & 60.24              & 84.14             \\
ClinicalBERT~\cite{huang2019clinicalbert} 
& 82.70              & 49.08              & 68.50              \\
ChestXRayBERT*~\cite{cai2021chestxraybert} 
& 83.12              & 49.55              & 69.19              \\
T5*~\cite{kale2020text}
& 83.90              & 49.68              & 75.82             \\
BlueBERT~\cite{bluebertpeng2020empirical}
& 84.00              & 48.44              & 68.71             \\
PubMedBERT~\cite{gao2021clip}
& 84.17              & 55.84              & 87.56             \\
BioELECTRA~\cite{raj2021bioelectra}
& 86.34              & 64.02              & 88.57             \\
SciFive~\cite{phan2021scifive}
& 86.50              & -                  & -                 \\ \hline
\textbf{Ours}                                         & \textbf{87.70}$_{\pm 0.8}$     & \textbf{66.12}$_{\pm 0.9}$     & \textbf{90.01}$_{\pm 0.8}$    \\ \hline
\end{tabular}}
	\label{text1}
\end{table}

\begin{table*}[!h]

	\centering
 \caption{Ablation study on the SLAKE, MIMIC-CXR(test-set) and the RSNA datasets. ``Prompt" represents the dynamic visual/textual prompts optimizing strategy. ``Multi-task" is the unified model jointly optimized with multiple pre-training tasks.}  
      \setlength{\tabcolsep}{3pt}
	
	\scalebox{0.85}{
\small
\begin{tabular}{ccccccccccccc}
\hline
\multirow{2}{*}{Prompt} & \multirow{2}{*}{Multi-task} & \multicolumn{3}{c}{SLAKE(\%)} & \multicolumn{5}{c}{MIMIC-CXR}                         & \multicolumn{3}{c}{RSNA Pneumonia(\%)} \\ \cline{3-13} 
                         &                             & Open   & Closed   & All   & BLEU-1 & BLEU-2 & BLEU-3 & BLEU-4 
                         & ROUGE-L & 1\%       & 10\%      & 100\%      \\ \hline
×                        & ×                           & 79.8$_{\pm 1.2}$    & 83.9$_{\pm 1.0}$      & 81.6$_{\pm 1.2}$   & 0.314$_{\pm 0.006}$   & 0.192$_{\pm 0.008}$   & 0.127$_{\pm 0.005}$   & 0.090$_{\pm 0.006}$    
& 0.265$_{\pm 0.004}$     & 87.5$_{\pm 1.1}$       & 88.2$_{\pm 0.9}$       & 89.5$_{\pm 1.2}$        \\
{\checkmark  }                        & ×                           & 83.7$_{\pm 1.4}$   & 88.5$_{\pm 1.3}$     & 85.5$_{\pm 1.1}$   & 0.381$_{\pm 0.008}$   & 0.286$_{\pm 0.007}$   & 0.184$_{\pm 0.004}$   & 0.133$_{\pm 0.007}$   
& 0.319$_{\pm 0.005}$     & 91.3$_{\pm 1.0}$       & 91.8$_{\pm 1.0}$       & 93.7$_{\pm 1.1}$        \\
×                        & {\checkmark  }                           & 83.1$_{\pm 1.1}$   & 87.5$_{\pm 1.2}$     & 84.6$_{\pm 1.2}$   & 0.379$_{\pm 0.006}$   & 0.247$_{\pm 0.007}$   & 0.174$_{\pm 0.005}$   & 0.120$_{\pm 0.005}$     & 0.307$_{\pm 0.007}$   
& 91.7$_{\pm 1.2}$       & 92.4$_{\pm 1.1}$       & 93.9$_{\pm 1.3}$        \\
\textbf{{\checkmark}}                       & \textbf{{\checkmark  }}                        & \textbf{84.0}$_{\pm 1.2}$    &\textbf{ 88.9}$_{\pm 0.9}$    & \textbf{85.9}$_{\pm 1.1}$   & \textbf{0.416}$_{\pm 0.005}$   & \textbf{0.271}$_{\pm 0.006}$   & \textbf{0.199}$_{\pm 0.006}$   & \textbf{0.148}$_{\pm 0.007}$   
& \textbf{0.335}$_{\pm 0.005}$     & \textbf{93.1}$_{\pm 1.2}$      & \textbf{94.6}$_{\pm 1.0}$     & \textbf{95.2}$_{\pm 1.1}$       \\ \hline

\end{tabular}}
	\label{ablation}
\end{table*}

\begin{table*}[h!]
	\centering
 \caption{Ablation study of various prompt methods which contains the static textual prompt, visual prompt and the proposed dynamic cross-modal prompts on the SLAKE, VQA-RAD, MELINDA datasets.}  
      \setlength{\tabcolsep}{7pt}
	\scalebox{0.85}{
\small
\begin{tabular}{cccccccccc}
\hline
\multicolumn{3}{c}{Prompt   strategy} & \multicolumn{3}{c}{SLAKE(\%)}                  & \multicolumn{3}{c}{VQA-RAD(\%)}             & MELINDA(\%)       \\ \hline
Textual Prompt          & Visual Prompt         & Dynamic Prompts        & Open          & Closed        & Overall       & Open          & Closed        & Overall & Accuracy           \\ \hline
×           & ×          & ×          & 81.0$_{\pm 1.3}$            & 85.2$_{\pm 1.2}$          & 82.5$_{\pm 1.3}$          & 66.4$_{\pm 1.2}$          & 82.5$_{\pm 1.1}$          & 77.1$_{\pm 1.2}$    & 77.8$_{\pm 1.2}$          \\
{\checkmark}           & ×          & ×          & 81.3$_{\pm 1.2}$          & 85.4$_{\pm 1.3}$          & 82.9$_{\pm 1.2}$          & 66.6$_{\pm 1.1}$          & 82.8$_{\pm 1.2}$          & 77.4$_{\pm 1.2}$    & 78.0$_{\pm 1.4}$          \\
×           & {\checkmark}           & ×          & 81.5$_{\pm 1.1}$          & 85.5$_{\pm 1.0}$          & 83.0$_{\pm 1.1}$          & 66.7$_{\pm 1.1}$          & 83.0$_{\pm 1.3}$          & 77.5$_{\pm 1.4}$    & 78.2$_{\pm 1.4}$          \\
{\checkmark}            & {\checkmark}           & ×          & 81.8$_{\pm 1.4}$          & 86.2$_{\pm 1.3}$          & 83.5$_{\pm 1.4}$          & 66.9$_{\pm 1.3}$          & 83.0$_{\pm 1.4}$          & 77.6$_{\pm 1.3}$    & 78.1$_{\pm 1.3}$          \\
\textbf{×}  & \textbf{×} & \textbf{{\checkmark}} & \textbf{84.0}$_{\pm 1.0}$ & \textbf{88.9}$_{\pm 1.1}$ & \textbf{85.9}$_{\pm 1.1}$ & \textbf{68.5}$_{\pm 1.4}$ & \textbf{84.4}$_{\pm 1.1}$ &\textbf{79.2}$_{\pm 1.3}$    & \textbf{79.6}$_{\pm 1.3}$ \\ \hline
\end{tabular}
}
	\label{ablation_prompt}
\end{table*}

\begin{table*}[h!]
	\centering
 \caption{Ablation study of multiple pre-training tasks across the VQA-RAD, IU X-Ray and CheXpert datastes.}  
      \setlength{\tabcolsep}{1pt}
	
	\scalebox{0.85}{
\small
\begin{tabular}{lcccccccccccc}
\hline
\multirow{2}{*}{Pre-training Tasks} & \multicolumn{3}{c}{VQA-RAD(\%)} & \multicolumn{6}{c}{IU X-Ray}                          & \multicolumn{3}{c}{CheXpert(\%)} \\ \cline{2-13} 
                                         & Open    & Closed   & All    & BLEU-1 & BLEU-2 & BLEU-3 & BLEU-4 & METEOR & ROUGE\_L & 1\%     & 10\%    & 100\%    \\ \hline
MLM                                      & 66.2$_{\pm 1.3}$     & 83.9$_{\pm 1.2}$      & 77.8$_{\pm 1.3}$    & 0.493$_{\pm 0.008}$  & 0.333$_{\pm 0.006}$  & 0.248$_{\pm 0.005}$  & 0.174$_{\pm 0.007}$  & 0.232$_{\pm 0.006}$  & 0.410$_{\pm 0.005}$     & 90.1$_{\pm 1.3}$    & 90.3$_{\pm 1.0}$    & 90.2$_{\pm 1.1}$     \\
MLM+ITM                                  & 67.1$_{\pm 1.1}$     & 84.0$_{\pm 1.0}$        & 78.3$_{\pm 1.2}$   & 0.509$_{\pm 0.006}$  & 0.335$_{\pm 0.005}$  & 0.252$_{\pm 0.004}$  & 0.182$_{\pm 0.005}$  & 0.239$_{\pm 0.006}$  & 0.421$_{\pm 0.006}$    & 90.2$_{\pm 1.1}$    & 90.3$_{\pm 1.1}$    & 91.4$_{\pm 1.2}$     \\
MLM+ITC                                  & 67.8$_{\pm 1.2}$    & 84.2$_{\pm 1.0}$     & 78.7$_{\pm 1.2}$   & 0.514$_{\pm 0.006}$  & 0.338$_{\pm 0.006}$  & 0.256$_{\pm 0.005}$  & 0.188$_{\pm 0.005}$  & 0.243$_{\pm 0.004}$  & 0.422$_{\pm 0.004}$    & 90.3$_{\pm 1.0}$    & 90.5$_{\pm 1.1}$    & 91.4$_{\pm 1.1}$     \\
\textbf{MLM+ITC+ITM  }                            & \textbf{68.1}$_{\pm 1.0}$    & \textbf{84.4}$_{\pm 1.1}$    & \textbf{79.0}$_{\pm 1.1}$     & \textbf{0.528}$_{\pm 0.007}$ & \textbf{0.343}$_{\pm 0.004}$  & \textbf{0.267}$_{\pm 0.004}$  & \textbf{0.196}$_{\pm 0.006}$  & \textbf{0.251}$_{\pm 0.005}$ & \textbf{0.423}$_{\pm 0.004}$    &\textbf{90.5}$_{\pm 1.2}$    & \textbf{90.8}$_{\pm 1.3}$   & \textbf{91.7}$_{\pm 1.0}$     \\ \hline
\end{tabular}}
	\label{ablation_task3}
\end{table*}

\begin{table}[h]
	\centering
 \caption{Ablation study of prompts. +P: pre-train with prompt. VP-N: select N visual prompts, TP-N:  select N textual prompts.}  
      \setlength{\tabcolsep}{2.pt}
	\abovedisplayskip=-0.8cm
	\scalebox{0.78}{
\small
\begin{tabular}{lcccccccccc}
\hline
Settings  & MLM & +P & ITM & ITM+P & VP-16 & -\textbf{49} & -64 & TP-16 & -\textbf{32} & -48 \\ \hline
SLAKE(Overall)   & 83.8  & \textbf{85.6}  & 83.3  & \textbf{85.3}  & 85.8  & \textbf{86.0}  & 85.6  & 85.9  & \textbf{86.1}  & 85.7  \\ 
CheXpert(100\%) & 88.2  & \textbf{90.3}  & 88.4  & \textbf{90.1}  & 91.2  & \textbf{91.6}  & 91.4  & 91.0  & \textbf{91.4}  & 91.3  \\ \hline
add-Memory(G)   & -  & 2.3           & -  & 1.4           & 0.9  & 1.3           & 2.2  & 0.4  & 0.5           & 0.7  \\
add-Param(M)     & -   & 0.88            & -   & 0.93           & 0.19   & 0.59            & 0.77   & 0.14  & 0.29            & 0.43   \\ \hline
\end{tabular}}
\captionsetup{font={small}}
	\label{p}
\end{table}
\begin{table}[h]
	\centering
 \caption{Ablation study of the size of prompt space M, J.}  
      \setlength{\tabcolsep}{4pt}
	\scalebox{0.78}{
\small
\begin{tabular}{cccccccccc}
\hline
M        & \multicolumn{3}{c}{512} & \multicolumn{3}{c}{\textbf{1024}} & \multicolumn{3}{c}{2048} \\ \hline
J        & 512    & 1024   & 2048  & 512    & \textbf{1024}   & 2048   & 512    & 1024   & 2048   \\ \hline
SLAKE(Overall)    & 84.7   & 84.9   & 84.8  & 85.0     & \textbf{85.6}   & 84.9   & 84.8   & 85.0     & 84.9   \\
CheXpert(100\%) & 91.0   & 91.2   & 91.1  & 91.3   & \textbf{91.7}   & 91.2   & 91.0   & 91.1   & 91.0   \\
MedNLI   & 87.4   & 87.6   & 87.5  & 87.4   & \textbf{87.7}   & 87.5   & 87.3   & 87.4   & 87.4   \\ \hline
\end{tabular}}
	\label{MJ}
\end{table}

\subsubsection{Medical Image Classification and Segmentation}
   We compare UniDCP with advanced works through labeling $1\%$, $10\%$ and $100\%$ ratios of CheXpert and RSNA Pneumonia datasets of classification and SIIM, RSNA of segmentation task in Table \ref{mic}.
   UniDCP outperforms SOTA works~\cite{zhou2023advancing,m3ae,tiu2022expertCheXzero,zhang2023knowledgeKAD,convirt,huang2021gloria,wu2023medklip,zhou2022generalized} under different label ratios over $4$ datasets.
   Specifically, when labeling a few datasets like 
     $1\%$ and $10\%$, UniDCP obtains $1.4\%$, $1.4\%$ and $1.8\%$, $1.2\%$ improvements compared with the SOTA work MPMA-M~\cite{zhang2023multi} on the CheXpert and RSNA datasets, and $6.4\%$, $1.1\%$ and $5.3\%$, $4.1\%$ compared with the SOTA work MRM~\cite{zhou2023advancing} on SIIM and RSNA respectively. This demonstrates the superior generalization of UniDCP with few labeled datasets. 
     Besides, for $100\%$ ratio, UniDCP has improved by $1.3\%$ and  $1.1\%$ on both datasets compared with the SOTA work MPMA~\cite{zhang2023multi}.
     In addition, under various data partitioning schemes of SIIM~\cite{siimAimoldin2019} (70\%/15\%/15\% for MRM$\dag$~\cite{zhou2023advancing}  and 70\%/30\%/30\% for MRM~\cite{zhou2023advancing} in Table~\ref{mic}), our model consistently demonstrates superior performance, 
     illustrating that the unified model benefits from shareable clinic knowledge.

            \begin{table}[h!]
	\centering
 \caption{Ablation study of pre-training datasets. -M: Pre-trained on the MIMIC-CXR, -M,R: Pre-trained on MIMIC-CXR and ROCO.}  
      \setlength{\tabcolsep}{1pt}
	\scalebox{0.85}{
\small
\begin{tabular}{llllllll}
\hline
\multicolumn{1}{c}{\multirow{2}{*}{Methods}} & \multicolumn{3}{c}{VQA-RAD}                                                           & \multicolumn{3}{c}{CheXpert}                                                   & MELINDA  \\ \cline{2-8} 
\multicolumn{1}{c}{}                         & \multicolumn{1}{c}{Open} & \multicolumn{1}{c}{Closed} & \multicolumn{1}{c}{Overall} & \multicolumn{1}{c}{1\%} & \multicolumn{1}{c}{10\%} & \multicolumn{1}{c}{100\%} & Accuracy \\ \hline
M3AE~\cite{m3ae}                        & 67.2                     & 83.5                       & 77.0                        & 86.2                    & 87.3                     & 87.9                      & 78.5     \\ \hline
\textbf{Ours-M}                      & \textbf{68.0}$_{\pm 1.3}$                     & \textbf{84.1}$_{\pm 1.2}$                       & \textbf{78.8}$_{\pm 1.3}$                        & \textbf{90.0}$_{\pm 0.9}$                    & \textbf{90.8}$_{\pm 0.7}$                     & \textbf{91.2}$_{\pm 0.8}$                      & \textbf{79.2}$_{\pm 0.6}$     \\
\textbf{Ours-M, R}                    & \textbf{68.5}$_{\pm 1.2}$                    & \textbf{84.4}$_{\pm 1.1}$                       & \textbf{79.2}$_{\pm 1.3}$                        & \textbf{90.5}$_{\pm 0.6}$                    & \textbf{91.2}$_{\pm 0.8}$                     & \textbf{91.9}$_{\pm 0.8}$                      & \textbf{79.6}$_{\pm 0.7}$     \\ \hline
\end{tabular}
}
	\label{ablation_vlp_data}
\end{table}
\begin{table}[h!]
	\centering
 \caption{Ablation study of lingual biomedical pre-training datasets. -O: Pre-trained on the original Med-VLP datasets. -P: Pre-trained on the biomedical domain-specific corpora.}  
      \setlength{\tabcolsep}{1pt}
	\scalebox{0.85}{
\small
\begin{tabular}{ccccc}
\hline
Methods    & Lingual Pre-train Data & MedNLI & PubMedQA & BioAsq \\ \hline
BioELECTRA~\cite{raj2021bioelectra} & PMC, PubMed          & 86.34  & 64.02    & 88.57  \\
SciFive~\cite{phan2021scifive}    & PMC, PubMed          & 86.50  & -        & -      \\ \hline
\textbf{Ours-O}    & -                   & \textbf{87.70}$_{\pm 1.3}$  & \textbf{66.12}$_{\pm 1.2}$    & \textbf{90.01}$_{\pm 1.3}$  \\
\textbf{Ours-P}    & PMC, PubMed          & \textbf{88.25}$_{\pm 1.2}$  & \textbf{66.98}$_{\pm 1.1}$    & \textbf{90.56}$_{\pm 1.2}$  \\ \hline
\end{tabular}
}
	\label{ablation_text_data}
\end{table}
\begin{figure*}[!htbp]
\centering
\includegraphics[width=1\linewidth]{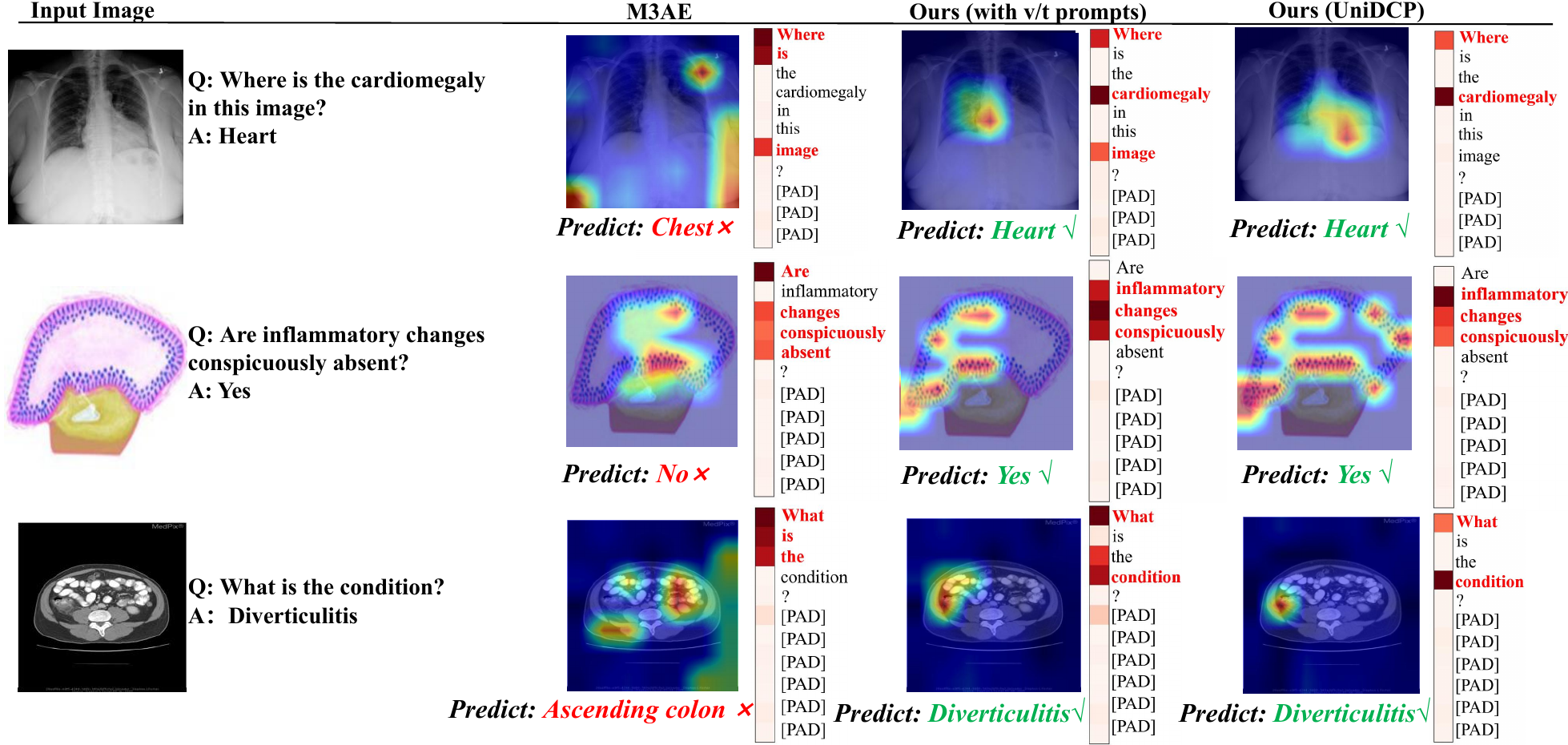}
\caption{The qualitative analysis is conducted on the Med-VQA tasks. First column: input image and question-answer pair. The second column to the last contains the cross-modal attention maps of the M3AE~\cite{m3ae}, the baseline equipped with the dynamic vision-language prompts and the proposed UniDCP model.}
\label{vqa2}
\end{figure*}
\begin{figure*}[h]
\centering
\includegraphics[width=1\linewidth]{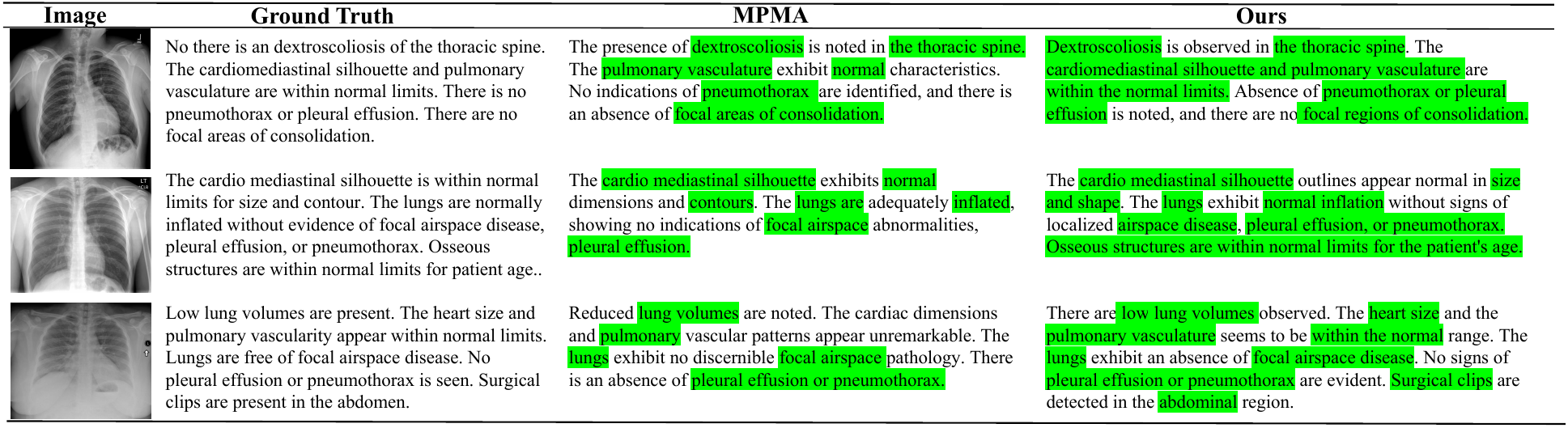}
\caption{The additional qualitative analysis is conducted on the medical report generation tasks across IU X-ray dataset. First column: Input medical image. Second column: the ground truth reports. Third column: the generated report of the SOTA model MPMA~\cite{zhang2023multi}. The last column: the generated report of ours.  The indication in Green: the correctly predicted MeSH terms.}
\label{medical report generation}
\end{figure*}
\subsubsection{Natural Language Inference and Question Answering}
As shown in Table \ref{text1}, we conduct comparisons on the medical natural language inference task over the MedNLI and question-answering task on PubMedQA and BioAsq. It should be noted that UniDCP achieves SOTA results compared with advanced BERT-based text-to-text methods like BioBERT~\cite{lee2020biobert}, ClinicalBERT~\cite{huang2019clinicalbert}, ChestXRayBERT~\cite{cai2021chestxraybert}, T5~\cite{kale2020text}, BlueBERT~\cite{bluebertpeng2020empirical}, PubMedBERT~\cite{gao2021clip}, BioELECTRA~\cite{raj2021bioelectra} and yields $87.70$\%, $66.12$\% and $90.01$\% accuracy on triple datasets. 
Explicitly, UniDCP considerably surpasses the SciFive~\cite{phan2021scifive} which is a domain-specific model pre-trained on large biomedical corpora by $1.20$\% on MedNLI. These demonstrate that UniDCP is highly versatile to medical text tasks with learned clinic knowledge.

\subsubsection{Medical Image-text Classification and Retrieval}
    We fine-tune UniDCP on the medical image-text classification and retrieval tasks. 
    As shown in Table \ref{cross-modal1}, UniDCP has significant improvement compared with the advanced works~\cite{liu2021contrastive,clipeslami2021does}. 
    In particular, although the M3AE enhances the visual-textual representations and yields $78.5\%$ accuracy, UniDCP improves the accuracy by $1.1\%$. This illustrates that UniDCP could grasp unified representations for precise classification. Besides, our model is superior to the advanced cross-modal pre-training works ViLT~\cite{kim2021vilt}, METER~\cite{dou2022empirical} and 
    considerably surpasses the M3AE by $3.35\%$, $5.35\%$ and $5.55\%$ with Recall@1, 5, 10 of text-to-image retrieval.
    This demonstrates that UniDCP has excellent plastically of medical vision-textual knowledge.

\begin{figure*}[!htbp]
\centering
\includegraphics[width=1\linewidth]{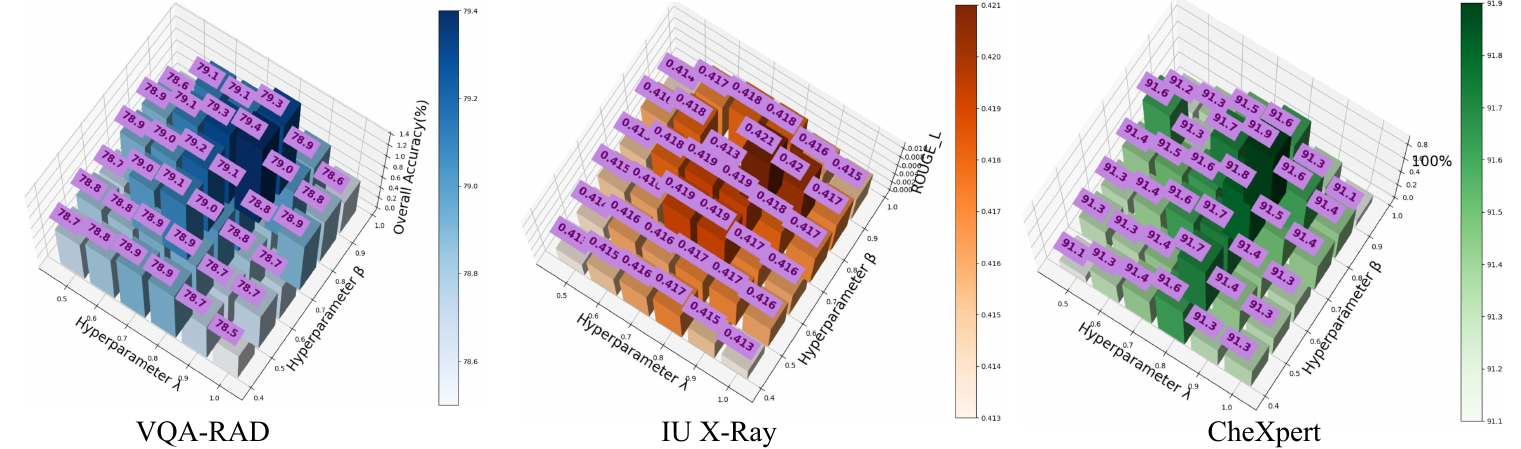}
\caption{The influence experiments of the loss balancing hyperparameters $\lambda$ and $\beta$ on the VQA-RAD, IU X-Ray and CheXpert datasets.}
\label{hyper}
\end{figure*}
\begin{figure*}[h]
\centering
\includegraphics[width=1\linewidth]{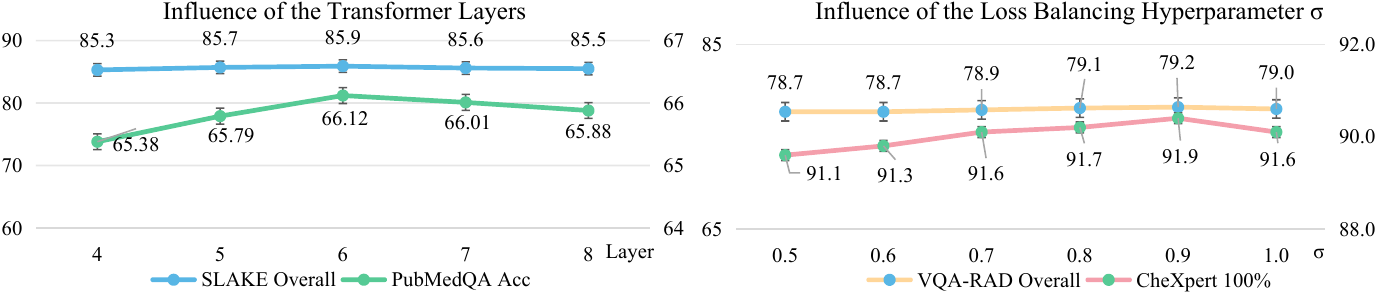}
\caption{Influence of the transformer layers in the unified model and the influence experiments of the loss balancing hyperparameters $\sigma$.}
\label{layer}
\end{figure*}
\subsection{Ablation Study}
\subsubsection{Ablation Study of the Proposed Methods}
As shown in Table \ref{ablation}, we conduct the ablation study to verify the effectiveness of the proposed methods, including the unified model optimized with multiple pre-training tasks and the dynamic cross-modal prompts optimizing strategy. For row $1$ in Table \ref{ablation}, we take the dual-encoder with uni-modal inputs and pre-train with a single MLM task as the baseline. For row $2$, the baseline equipped with dynamic prompts optimizing strategy obtains $3.9$\% overall accuracy on SLAKE, $0.043$ and $0.064$ of BLEU-4 and ROUGE-L on MIMIC-CXR, $3.8$\%, $3.6$\% and $4.2$\% of the three different labeled ratios on RSNA Pneumonia. 
This demonstrates that the dynamic cross-modal prompts unify the vision-language inputs and share the clinic knowledge for better understanding, bringing obvious improvements in medical uni-modal and cross-modal tasks.

For row $3$ in Table \ref{ablation}, the baseline is adapted to the unified model pre-trained with MLM, ITM and ITC tasks, obtaining $3.0$\% improvements on SLAKE, $0.030$ and $0.042$ of BLEU-4 and ROUGE-L on MIMIC-CXR, $4.2\%$, $4.2\% $ and $4.4$\% on RSNA Pneumonia. It illustrates the unified model with multiple pre-training tasks benefits downstream tasks with shareable knowledge. Moreover, we also take a further step to demonstrate the effectiveness of each pre-training task in Table~\ref{ablation_vlp_data} and figure out that the ITC task obverses the most significant improvements for the alignment. 

Accordingly, the unified baseline jointly equipped with the dynamic prompts optimizing strategy and multiple pre-training tasks obverses the best result which reaches 
$85.9$\% overall accuracy on SLAKE, $0.335$ of ROUGE-L on MIMIC-CXR, $93.1\%$, $94.6\% $ and $95.2$\% on RSNA Pneumonia, suggesting that our proposed model grasps efficient clinic knowledge and the unified cross-modal representations and then adapt to multiple fine-tuning tasks flexibly.

\subsubsection{Ablation Study of Dynamic Prompts Strategy}

We offer an ablation study of distinct prompt methodologies, encompassing static visual/textual prompts and the novel dynamic prompt strategy, as depicted in Table \ref{ablation_prompt}. The experiments unveil that static visual and textual prompts yield minimal enhancements across multiple medical fine-tuning tasks. Conversely, the introduced dynamic cross-modal prompts exhibit substantial superiority across both tasks. 
This demonstrates the advantage of dynamic prompts in integrating shared medical information across multiple downstream tasks.

\subsubsection{Ablation Study of Pre-training Tasks}
\label{F}
We conduct the ablation study of multiple pre-training tasks within a unified model in Table~\ref{ablation_task3}.  The experiments on the VQA-RAD, IU X-Ray and CheXpert datasets illustrate the effectiveness of each pre-training task and the ITC task obverses the most significant improvement.

\subsubsection{Ablation Study of Pre-training Datasets}
\label{G}
Table~\ref{ablation_vlp_data} illustrates the ablation study of Med-VLP pre-training datasets. It reveals that while pre-training UniDCP with a single pre-training dataset may lead to a decrease in performance, our framework still offers advantages compared to state-of-the-art method M3AE~\cite{m3ae} across the SLAKE, CheXpert, and MELINDA datasets. This is attributed to the incorporation of the unified structure jointly with the dynamic prompt optimizing strategy to incorporate clinic knowledge into multiple medical uni-modal and cross-modal tasks.

\subsubsection{Lingual Biomedical Pre-training Datasets}

As shown in Table~\ref{ablation_text_data}, the ablation study of the lingual biomedical pre-training datasets is adopted in the medical natural language inference task and the medical question answering task. From the ablation study, it is evident that even without specialized pre-training on medical text datasets like PubMed abstracts (PubMed)~\cite{jin2019pubmedqa} and PubMed Central full-text articles (PMC)~\cite{zhang2023pmc}, our model can achieve remarkable performance compared with the SOTA methods like BioELECTRA~\cite{raj2021bioelectra} and SciFive~\cite{phan2021scifive}. Furthermore, when we utilize the same  Lingual biomedical datasets PMC and PubMed for pre-training, more pronounced improvements are attainable across the MedNLI, PubMedQA and BioAqa datasets.

\subsubsection{Addition Memory and Parameter
Cost}
We conducted the ablation study to examine how the inclusion of different numbers of learnable visual/textual prompts affects additional memory and parameter costs, as detailed in Table~\ref{p}}. Besides, Table~\ref{MJ} illustrates the ablation study of $M$, $J$ in Eq.~\ref{mn}.
 These reveal that the dynamic prompts achieve minimal memory cost, and more importantly, {the selection of effective prompts} is more useful than the construction of more of them.

\subsection{Qualitative Analysis}
As shown in Fig.~\ref{vqa2}, we conduct a detailed qualitative analysis through the Grad-CAM to figure out how a unified model with dynamic prompts helps medical vision-language prediction. Besides, 
we also demonstrate the model's superior ability to generate accurate and semantically meaningful medical reports in Fig.~\ref{medical report generation}.

Specifically, for the abdominal CT selected from VQA-RAD~\cite{vqaradlau2018dataset}, the proposed UniDCP can precisely indicate the keywords ``What, condition" with the precise visual object rather than the meaningless words ``is the" and the irrelevant intestinal parts (e.g. the black background) the M3AE~\cite{m3ae} focuses on.  Moreover, we can witness the prominent gains in the unification of the vision/text features from the second column to the fourth, the highlighted vision objects and keywords are closely semantically related and conduct the correct prediction in a unified manner.
The comparison demonstrates that UniDCP exhibits exceptional unification capabilities and effectively identifies precise clinical vision and textual features to facilitate unified cross-modal representations.

Besides, while the baseline equipped with dynamic vision-language prompts suffices to highlight the relatively coarse parts of black diverticulitis and the keywords with lightweight, UniDCP distinctly centers on a more specific visual object of diverticulitis and keywords with more distinct weights for indication.
It demonstrates the effectiveness of the dynamic prompts for learning the shareable clinic information and further illustrates 
importance of the unified model for more precise cross-modal representations.

\subsection{Influence Analysis}
\subsubsection{Influence of Layers Within Unified Model}
\label{C}
As shown in Fig. \ref{layer}, we also conduct the influence experiment on the transformer layer of the unified model across SLAKE~\cite{slakeliu2021slake}, RSNA~\cite{rsnashih2019augmenting} and PubMedQA~\cite{jin2019pubmedqa}. It can be witnessed that the unified model composed of $6$ transformer layers obtains the best results on the medical uni-modal and cross-modal tasks.

\subsubsection{Influence of Hyperparameters}
\label{D}
In this section, we conduct the influence experiment on the VQA-RAD, IU X-Ray, and CheXpert datasets of different hyperparameters illustrated in Fig. \ref{hyper} and Fig. \ref{layer}. We assign the hyperparameters $\sigma$, $\lambda$ and $\beta$ from $0.5$ to $1.0$ with step 0.1 and attach the best result when $\sigma$ is set to $0.9$, $\lambda$ is $0.8$, and $\beta$ is $0.9$. The stable experiment results illustrate the generalization and robustness of our proposed method.

\section{Conclusion}
In this paper, we construct a unified and plastic Med-VLP model for multiple medical vision-language tasks.
Our approach first dynamically selects visual/textual prompts within the shareable prompts space for harmonizing diverse inputs. The joint optimization of the dynamic prompts and the unified model alongside multiple pre-training tasks, facilitates the alignment of cross-modal representations while adapting the wealth of shareable information to multiple fine-tuning tasks.
The proposed framework exhibits superior performance on 
$8$ medical vision-language tasks over $14$ corresponding datasets, illustrating the outstanding scalability of UniDCP.
In future exploration, we plan to construct a novel
generative model that can handle more combinations of medical modalities. Codes will be released.



\bibliographystyle{IEEEtran}

\bibliography{ref.bib}

\newpage






\vfill

\end{document}